\documentclass[11pt,a4paper]{article}
\usepackage[hyperref]{emnlp2018}
\usepackage{times}
\usepackage{latexsym}

\aclfinalcopy
\usepackage{times}
\usepackage{color}
\usepackage{amsfonts}
\usepackage{graphicx}
\usepackage{amsmath,amsfonts,amssymb,bm}
\usepackage{booktabs} 
\usepackage{array}
\usepackage{boxedminipage}
\usepackage{paralist}
\usepackage{url}
\usepackage{diagbox}
\usepackage{enumitem}
\usepackage{paralist}
\definecolor{darkspringgreen}{rgb}{0.05, 0.5, 0.06}
\usepackage{arydshln}

\usepackage{microtype}      % microtypography
\usepackage{comment}        % bulk comment
\usepackage{amsopn}
\usepackage{xcolor}
\usepackage{multirow}
\usepackage{subcaption}

\usepackage{tabularx}
\usepackage{tikz,pgfplots}
\pgfplotsset{compat=1.14}

\definecolor{g-green}{HTML}{0F9D58}

\def\emnlppaperid{1301}

\newcommand{\red}[1]{\textcolor{red}{#1}}

\usepackage{url}

\newcommand{\blue}[1]{\textcolor{blue}{#1}}

\newcommand{\sys}{{\sc SciIE}}
\newcommand{\dataname}{{\sc SciERC}}
\newcommand{\mb}{{E2E Rel}}
\newcommand{\elmo}{{\sc ELMo}}
\newcommand{\coref}{{E2E Coref}}
\newcommand{\pipeline}{{Pipeline}}

\newcommand{\semevalEnt}{{SemEval 17}}
\newcommand{\semevalRel}{{SemEval 18}}

\newcolumntype{L}[1]{>{\centering\let\newline\\\arraybackslash\hspace{0pt}}m{#1}}
\newcolumntype{C}[1]{>{\centering\let\newline\\\arraybackslash\hspace{0pt}}m{#1}}
\newcolumntype{R}[1]{>{\centering\let\newline\\\arraybackslash\hspace{0pt}}m{#1}}

\title{Multi-Task Identification of Entities, Relations, and Coreference  
\\ for  Scientific Knowledge Graph Construction}
\author{
  Yi Luan\quad Luheng He \quad Mari Ostendorf \quad Hannaneh Hajishirzi \\
University of Washington\\
   {\{luanyi, luheng, ostendor, hannaneh\}@uw.edu}
}

\date{}

\begin{document}

\maketitle

\newcommand\luheng[1]{[\textcolor{orange}{LH: {#1}}]}

\begin{abstract}

%In this paper
We introduce a multi-task setup of identifying and classifying entities, relations, and coreference clusters in scientific articles. 
We create \dataname, a dataset that includes annotations for all three tasks 
and develop a unified framework called Scientific Information Extractor (\sys)  for  with shared span representations. 
The multi-task setup reduces cascading errors between tasks and leverages cross-sentence relations through coreference links.
Experiments show that our multi-task model outperforms previous models in scientific information extraction without using any domain-specific features. 
We further show that the framework supports
%can be extended to 
construction of a scientific knowledge graph, which we use to analyze information in scientific literature.\footnote{Data and code are publicly available at:  \url{http://nlp.cs.washington.edu/sciIE/}}

\end{abstract}

\section{Introduction}
As scientific communities grow and evolve, new tasks, methods, and datasets are introduced and different methods are compared with each other.
%known methods are used for new tasks; different methods are compared with each other; extensions of methods and tasks are introduced; etc. 
Despite advances in search engines, it is still hard to identify new technologies and their relationships with what existed before. 
To help researchers more quickly identify opportunities for new combinations of tasks, methods and data,
it is important to design intelligent algorithms that can extract and organize scientific information from a large collection of documents. 

\begin{figure}[t]
\centering
\includegraphics[width=\columnwidth]{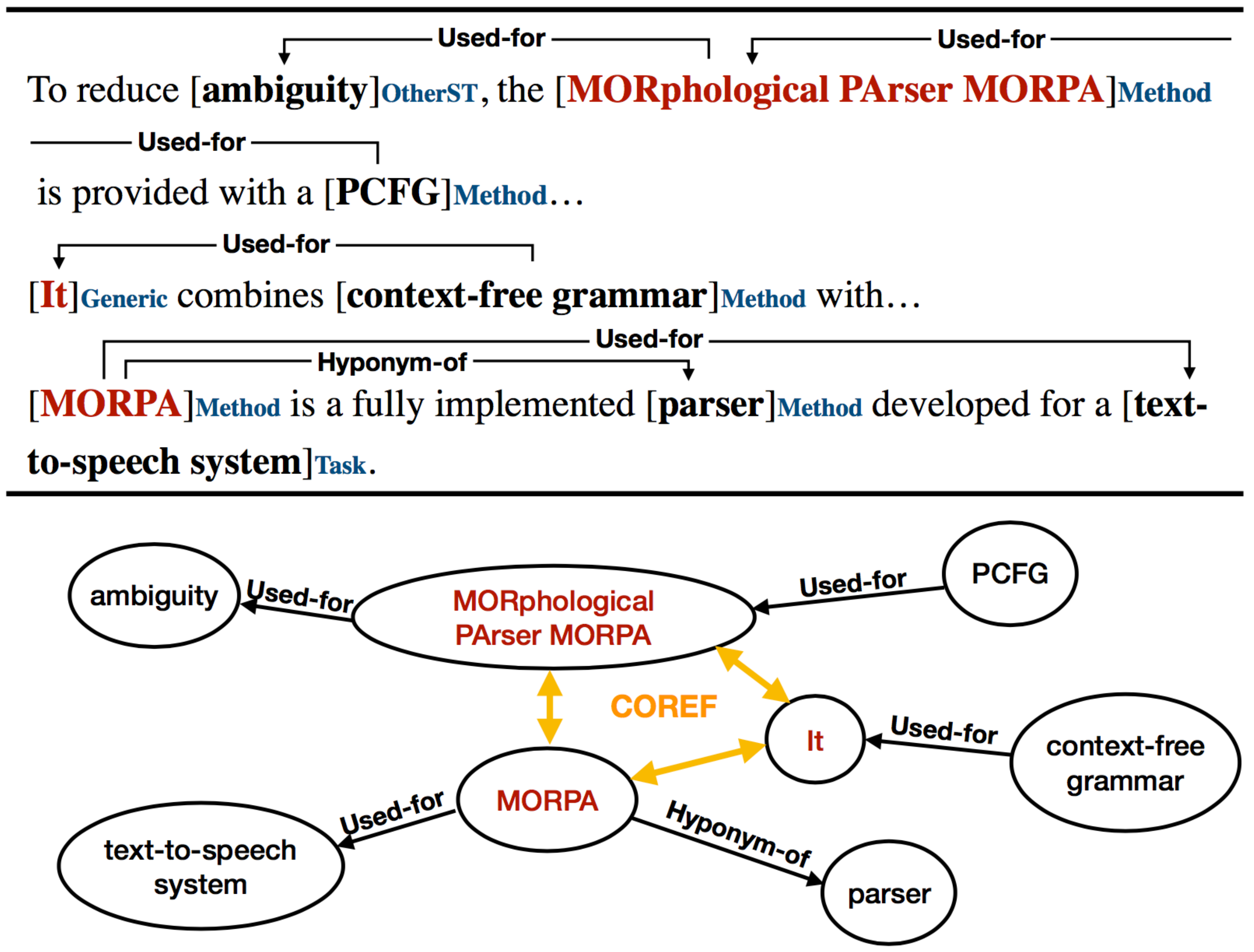}
% \vspace{-1em}
\caption{
Example annotation: phrases that refer to the same scientific concept are annotated into the same coreference cluster, such as  \textit{MORphological PAser MORPA}, \textit{it} and \textit{MORPA} (marked as red).  
}\label{fig:annotation}
% \vspace{-1em}
\end{figure}

Organizing scientific information into structured knowledge bases requires information extraction (IE) about scientific entities and their relationships. 
%(Figure \ref{fig:annotation}). 
However, the challenges associated with scientific IE are greater than for a general domain.  First, annotation of scientific text requires domain expertise which makes annotation costly and limits resources. In addition,  most relation extraction systems are designed for within-sentence relations. However,  extracting information from scientific articles requires extracting relations across sentences.
Figure~\ref{fig:annotation} illustrates this problem. The cross-sentence relations between some entities can only be connected by entities that refer to the same scientific concept, including generic terms (such as the pronoun \textit{it}, or phrases like \textit{our method}) that are not informative by themselves. With co-reference, \textit{context-free grammar} can be connected to \textit{MORPA} through the intermediate co-referred pronoun \textit{it}. 
Applying existing IE systems to this data, without co-reference, will result in much lower relation coverage (and a sparse knowledge base).

% Recent work has focused on extracting scientific terms in a sentence. Most recently, a new task is introduced for identifying scientific relations given pairs of entities in a sentence~\cite{x}. However, these two tasks are considered in isolation and are only focused Previous work in scientific information extraction has focused on sentence-level information extraction.  two individual tasks of scientific  These tasks are considered in isolation, and whithin-sentence and in isolation, while extracting information from scientific articles requires a document-level analysis of scientific terms and their relations across sentences.

%Previous work in scientific information extraction has focused on two isolated tasks of term extraction ~\cite{luan2017scienceie,gupta2011analyzing,tsai2013concept} and relation identification~\cite{SemEval2018Task7} within a sentence. 
In this paper, we 
%provide a unified %training and testing platform and 
%model for extracting scientific entities and their relations across sentences. In order to capture cross-sentence relations, we augment relation extraction with coreference resolution.
%and argue that these two tasks should be considered jointly. 
%More specifically, we identify generic terms (such as ``method'') and add coreference links between different entities in the document.  
%
develop a unified learning  model for extracting scientific entities, relations, and coreference resolution.
This is different from previous work~\cite{luan2017scienceie,gupta2011analyzing,tsai2013concept,SemEval2018Task7} which often addresses these tasks as independent components of a pipeline. Our unified model is a multi-task setup that  shares  parameters across low-level tasks, making predictions by leveraging context across the document through coreference links. 
Specifically, we extend prior work for learning span representations and coreference resolution \cite{Lee2017EndtoendNC,he2018jointly}.
Different from a standard tagging system, our system enumerates all possible spans during decoding and can effectively detect overlapped spans. 
It 
%, and can effectively 
avoids cascading errors between tasks by jointly modeling all spans and span-span relations. 

%By jointly modeling spans and leveraging cross-sentence information, our model can effectively improve performance across all tasks. 

To explore this problem, we create a dataset \dataname\ for scientific information extraction, which includes annotations of scientific terms, relation categories and co-reference links. 
%We randomly select 500 paper abstracts from four artificial intelligence communities (Computer Vision, Natural Language Process, Speech and Machine Learning) and ask domain experts to operate careful annotation.
Our experiments show that the unified model is better at predicting  span boundaries, and it outperforms previous state-of-the-art scientific IE systems on entity  and relation extraction \cite{luan2017scienceie,scienceIE}.
In addition, we
build a scientific knowledge graph integrating terms and relations extracted from each article.  
%Constructing this knowledge graph takes advantage of our unified model in two ways. Propagating scientific terms through coreference links can help extracting more relation triples. Moreover, general terms are disambiguated into more specific terms \ylcomment{the parsing /detection example}. \ylcomment{newly added:} 
Human evaluation shows that propagating coreference can significantly improve the quality of the automatic constructed knowledge graph.

%%MO: Taking the numbers out saves a line
In summary we make the following contributions. We create a dataset for  scientific information extraction by jointly annotating scientific entities, relations, and coreference links.  Extending a previous end-to-end coreference resolution system,  we develop a multi-task learning framework that can detect scientific entities, relations, and coreference clusters without hand-engineered features. We use our unified framework to build a scientific knowledge graph from a large collection of documents and analyze information in  scientific literature. 

% \begin{figure*}
%     \centering
%     \begin{subfigure}[b]{7.5cm}%{0.4\textwidth}
%         \includegraphics[width=\textwidth]{./figures/annotation_example.pdf}
%         \caption{\small{Document level knowledge graph built without coreference connection}}
%         \label{fig:KG_nocoref}
%     \end{subfigure}
%     ~~~~~~~~~ %add desired spacing between images, e. g. ~, \quad, \qquad, \hfill etc. 
%       %(or a blank line to force the subfigure onto a new line)
%     \begin{subfigure}[b]{7cm}%{0.4\textwidth}
%         \includegraphics[width=\textwidth]{./figures/KG_coref.pdf}
%         \caption{\small{Document level knowledge graph built with coreference connection.}
%         \label{fig:KG_coref}}
%     \end{subfigure}

%     \caption{\small{Knowledge graph constructed from our proposed annotation scheme for a scientific paper. The blue nodes indicates scientific terms, and the yellow nodes indicate the generic terms that are used for connection. Black arches indicate in-sentence relations and yellow arches indicate cross-sentence coreference relations. The graph built with coreference connections has more relation coverage than the one without coreference.}}
%     \label{fig:KG}
% \end{figure*}

%%MO: There are a lot of papers in here that should not have been cited because they are totally unrelated. I've taken them out.

\section{Related Work}
There has been growing interest in research on automatic methods for information extraction from scientific articles. Past research in scientific IE addressed analyzing citations~\cite{athar2012detection,athar2012context,kas2011structures,GBOR16.870, sim2012discovering,do2013extracting,jaidka2014computational,abu2011coherent}, analyzing research community~\cite{vogel2012he,anderson2012towards}, and unsupervised methods for  extracting scientific entities and relations~\cite{gupta2011analyzing, tsai2013concept,gabor2016unsupervised}. 

More recently, two datasets in SemEval 2017 and 2018 have been introduced, which facilitate research on supervised and semi-supervised learning for scientific information extraction.  \semevalEnt~\cite{scienceIE} includes 500 paragraphs from articles %in  ScienceDirect\footnote{www.sciencedirect.com}
in the domains of computer science, physics, and material science. It includes three  types of entities (called keyphrases): Tasks, Methods, and Materials and two relation types: hyponym-of and synonym-of. %The dataset includes a large portion of overlapped entities. 
\semevalRel~\cite{SemEval2018Task7} is focused on predicting relations between entities within a sentence. It consists of %\hanna{how many entity types?} and 
%%MO: no entity types
six relation types.  Using these datasets, neural models~\cite{ammar2017ai2,ammar2018construction, luan2017scienceie,augenstein2017multi} are introduced for extracting scientific information. %show that semi-supervised augenstein2017multi} introduce neural models for entity extraction in scientific articles and show that  using semi-supervised learning and multi-task learning to neural based models to leverage large unannotated scholarly datasets for a scientific term extraction task \cite{augenstein2017multi}. \ylcomment{should we mention semeval 2017 2018 in here or the data sectio n is enough}
We extend these datasets by increasing relation coverage, adding cross-sentence coreference linking, and removing some annotation constraints. 
%More details are described in the dataset section. 
Different from most previous IE systems for scientific literature and general domains~\cite{miwa2016end,xu2016improved,peng2017cross,quirk2016distant,luan2018uwnlp,adel2017global}, which use preprocessed syntactic, discourse or coreference features as input, 
our unified framework does not rely on any pipeline processing and is able to model overlapping spans. 

While \newcite{Singh2013JointIO}  show improvements by jointly modeling entities, relations, and coreference links, 
most recent neural models for these  tasks  focus on single tasks \cite{Clark2016ImprovingCR,Wiseman2016LearningGF,Lee2017EndtoendNC,lample2016neural,peng2017cross} or joint entity  and relation extraction \cite{katiyar2017going,zhang2017end,adel2017global,zheng2017joint}. Among those studies, many papers assume the entity boundaries are given, such as \cite{Clark2016ImprovingCR}, \newcite{adel2017global} and \newcite{peng2017cross}.
 Our work relaxes this constraint and predicts entity boundaries by
 %%MOedit
 %enumerating 
 optimizing over all possible spans.
Our model draws  from recent end-to-end span-based models for coreference resolution \cite{Lee2017EndtoendNC,lee2018higher} and semantic role labeling \cite{he2018jointly} and extends them for the multi-task framework involving the three tasks of identification of entity, relation and coreference. 

Neural multi-task learning has been applied to a range of NLP tasks. Most of these models share word-level representations \cite{collobert2008unified,Klerke2016ImprovingSC,luan2016multiplicative,luan2017multi,Rei2017SemisupervisedML}, while \newcite{peng2017cross} uses high-order cross-task factors. Our model instead propagates cross-task information via span representations, which is related to \newcite{Swayamdipta2017FrameSemanticPW}.

%Neural tagging models have been recently introduced to tagging problems such as NER. For example,  \citet{collobert2011natural} use a CNN over a sequence of word embeddings and apply a CRF layer on top.  \citet{huang2015bidirectional} use hand-crafted features with LSTMs to improve performance. There is currently great interest in using character-based embeddings in neural models.  \cite{chiu2015named,lample2016neural,ballesteros2015improved,ma2016end}. Our approach also takes advantage of neural tagging models and character-based embeddings for IE in scientific articles. 

% \rev{ Previous work on semi-supervised learning for neural models has mainly focused on transfer learning \cite{dai2015semi,luan2014semi,harsham2015driver} or initializing the model with pre-trained word embeddings \cite{mikolov2013efficient,pennington2014glove,levy2014dependency,luan2016lstm,luan2015efficient,luan2016multiplicative}. In our work, we use pre-training but also use more powerful methods including graph-based semi-supervision \cite{subramanya2011semi,liu2013graph,liu2015acoustic,liu2016graph,liu2016novel} and a method for leveraging partially labeled data \cite{kim2015weakly}. We show that the combination of these techniques gives better results than any one alone.}

\section{Dataset}
\label{sec:dataset}
Our dataset (called \dataname)  includes  annotations for scientific entities, their relations, and coreference clusters for 500 scientific abstracts. These abstracts are taken from 12 AI conference/workshop proceedings in four AI communities from the Semantic Scholar Corpus\footnote{These conferences include  general AI~(AAAI, IJCAI), NLP~(ACL, EMNLP, IJCNLP), speech~(ICASSP, Interspeech), machine learning~(NIPS, ICML), and computer vision~(CVPR, ICCV, ECCV) at \url{http://labs.semanticscholar.org/corpus/}}.
\dataname\ extends previous datasets in scientific articles SemEval 2017 Task 10 (\semevalEnt)~\cite{scienceIE} and SemEval 2018 Task 7 (\semevalRel)~\cite{SemEval2018Task7} by extending entity types, relation types, relation coverage, and adding cross-sentence relations using coreference links.  Our dataset is publicly available at:  \url{http://nlp.cs.washington.edu/sciIE/}.
Table~\ref{tab:data} shows the statistics of \dataname.

% \vspace{-.2cm}
\paragraph{Annotation Scheme}
We define  six types for annotating scientific entities (Task, Method, Metric, Material, Other-ScientificTerm and Generic) and seven relation types (Compare, Part-of, Conjunction, Evaluate-for, Feature-of, Used-for, Hyponym-Of).  Directionality is taken into account except for the two symmetric relation types (Conjunction and Compare). 
Coreference links are annotated between identical scientific entities. 
A Generic entity is annotated only when  the entity is involved in a relation or is coreferred with another entity.  
Annotation guidelines can be found in Appendix~\ref{sec:annotation}.
Figure~\ref{fig:annotation} shows an annotated example. 

Following annotation guidelines from~\newcite{qasemizadeh2016acl} and using the BRAT interface \cite{stenetorp2012brat}, our annotators perform a greedy annotation for spans and always prefer the longer span whenever ambiguity occurs.  Nested spans are allowed when a subspan has a relation/coreference link with another term outside the span. 
%\hanna{we need to mention how long did it take to annotate every document on average}

\begin{table}[t]
\newcolumntype{Y}{>{\centering\arraybackslash}X}
\newcommand{\colindent}{\;}
\setlength{\tabcolsep}{.25em}
\footnotesize
\centering
\begin{tabularx}{\linewidth}{l c *{3}{Y}}
    \toprule
     Statistics & \dataname & \semevalEnt & \semevalRel \\    
    \midrule
%    \#Docs   & 500 & 500 & 500\\
%   \#Entity types & 6 & 3 & -  \\
%   \#Relation types   & 7 & 2 & 6\\
   \#Entities  & 8089 & 9946 & 7483 \\
   \#Relations  & 4716 & 672 & 1595  \\
    \#Relations/Doc  & 9.4 & 1.3 & 3.2  \\
   \#Coref links& 2752 & - & -  \\
   \#Coref clusters& 1023 & - & -  \\
  \bottomrule
\end{tabularx}
\vspace{-1em}
\caption{Dataset statistics for our dataset \dataname\ and two previous datasets on scientific information extraction. All datasets annotate 500 documents.} %Comparison between other two information extraction tasks on scientific literature. Our dataset covers more tasks than previous ones. Our relation coverage is significantly higher than that of SemEval 2018.} 
  \label{tab:data}
  \vspace{-1em}
\end{table}
% \begin{table}
%   \centering
%  {\footnotesize
%   \begin{tabular}{l|lll}
%     \toprule
%      Model & Ours & SemEval 2017 & SemEval 2018 \\    
%     \midrule
%   \#sentence & -& - & -  \\
%   \#Keyphrases  & - & - & - \\
%  \#relation   &- & - & -  \\
%   \#coreference    &- & - & -  \\

%     \bottomrule
%   \end{tabular}}
%   \caption{Comparison between other two information extraction tasks on scientific literature. Our system covers most tasks. Our relation coverage is significantly higher than SemEval 2018} 
%   \label{tab:data}
% \end{table}

% \vspace{-.2cm}
\paragraph{Human Agreements}
One domain expert annotated all the documents in the dataset; 12\% of the data is dually annotated by 4 other domain experts to evaluate the user agreements. %four other domain experts annotate \hanna{xx percent} amount of documents (between 4 to 7 documents per person) in order to calculate human agreement.  
%Each annotator was shown with the span that has been annotated by the main annotator, and was asked to correct some obvious span errors as well as annotating keyphrase types, links and types (if any) of relation and coreference. 
The kappa score for annotating entities is 76.9\%,  relation extraction is 67.8\% and  coreference is 63.8\%. %\luheng{need to include details if we have space}

% \vspace{-.2cm}
\paragraph{Comparison with previous datasets} 
\dataname\ is focused on annotating cross-sentence relations and has more relation coverage than \semevalEnt\ and \semevalRel, as shown in Table \ref{tab:data}. 
\semevalEnt\ is mostly designed for entity recognition and only covers two relation types. The task in \semevalRel\ is to classify a relation between a pair of entities given entity boundaries,
%There are two constraints in annotating relations in \semevalRel: the relations are only annotated within a sentence, and each entity can only appear in one relation. %
but only intra-sentence relations are annotated and each entity only appears in one relation, resulting in sparser relation coverage than our dataset (3.2 vs.\ 9.4 relations per abstract). 
%Therefore,  \semevalRel\ has sparse relation coverage (3.2 relations per abstract), whereas our dataset has 9.4 relation coverage per abstract. 
\dataname\ extends these datasets by adding more relation types and coreference clusters, which allows representing cross-sentence relations, and removing annotation constraints. Table~\ref{tab:data} gives a comparison of statistics among the three datasets.
%With the dataset we are able to build a knowledge graph on AI communities and are able to perform meaningful analysis on it.
%\luheng{the order of sentences is a bit weird}
%\hanna{Yi: check if what I say is accurate}
In addition, \dataname\ aims at including broader coverage of general AI communities.
% \footnote{Some documents are shared between our dataset and \semevalRel; we annotate based on their gold entity boundaries for those documents.}  

\begin{figure*}[t]
\centering
\includegraphics[width=0.9\textwidth, keepaspectratio,trim={0cm 4.8cm 0cm 6.6cm},clip]{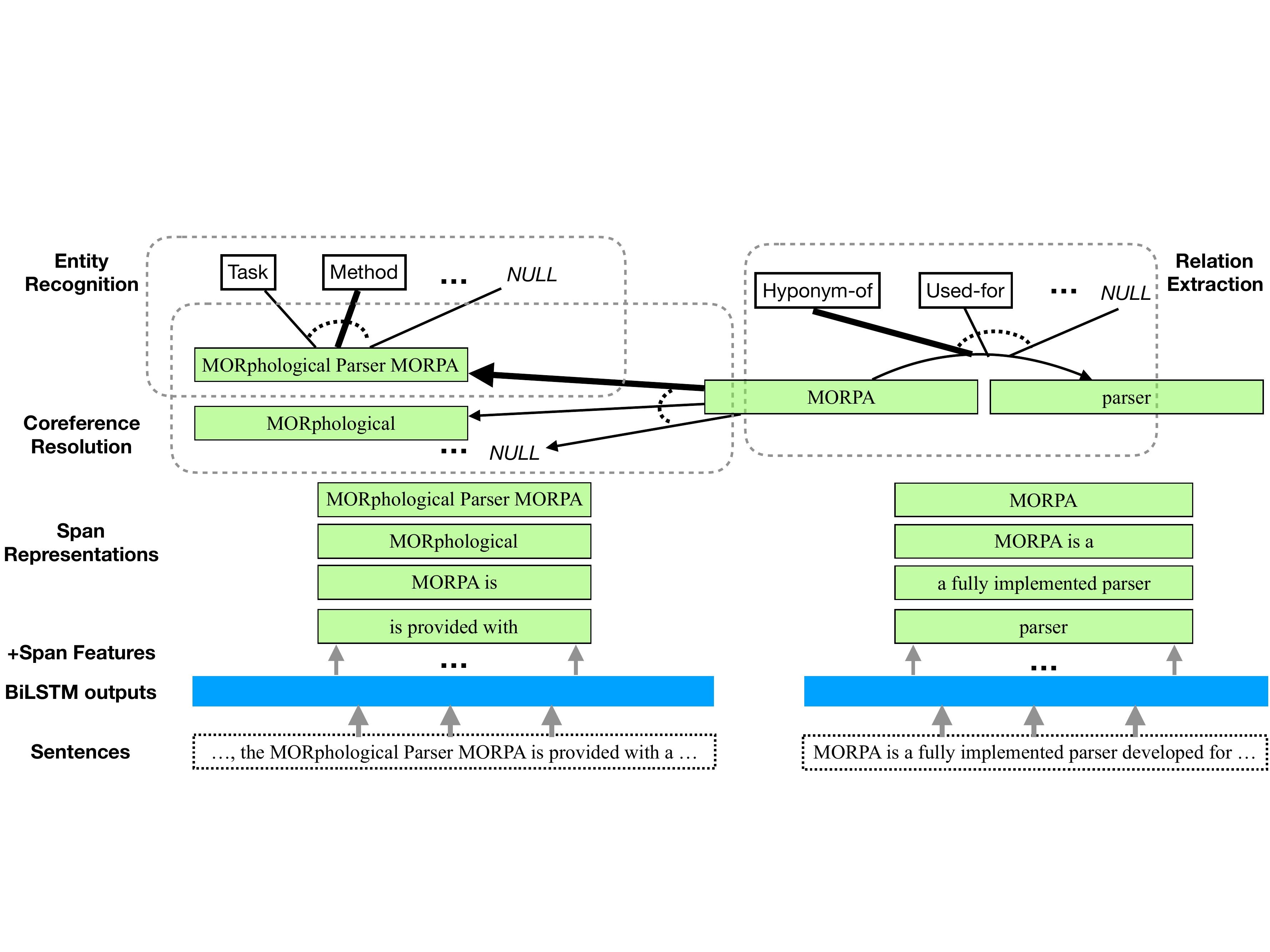}
% \vspace{-1em}
\caption{
Overview of the multitask setup, where all
three tasks are treated as classification problems on top of shared span representations. Dotted arcs indicate the normalization space for each task.
%%MO: put sentence below back if we have room
%The Coreference Resolution task propagates cross-sentence information via the vector representation for \emph{MORPA} and benefits the other two sentence-level tasks. 
%\hanna{suggestion: what if we move span phrases into the green bars (and then probably use a bit lighter color)? The figures has  a lot of text, a bit hard to read} 
}\label{fig:model:mtl}
% \vspace{-1em}
\end{figure*}

\section{Model}
\label{sec:model}
We develop a unified framework (called \sys) to identify and classify scientific entities, relations, and coreference resolution across sentences. \sys\ is a multi-task learning setup that extends  previous span-based models for coreference resolution \cite{Lee2017EndtoendNC} and semantic role labeling \cite{he2018jointly}. All three tasks of entity recognition, relation extraction, and coreference resolution 
%\luheng{not sure if I should keep them capitalized?} \hanna{No need to capitalize. let's call them Entity recognition, Relation extraction, and Coreference resolution}
are treated as multinomial classification problems with shared span representations. 
%For Relation Extraction, our model decides which relation type holds for each pair of spans, or there is none. For Entity Recognition, the model classifies the entity type (also including the none-type) for each span. 
\sys\ benefits from expressive contextualized span representations as classifier features. By sharing span representations, sentence-level tasks can benefit from  information propagated from coreference resolution across sentences, without increasing the complexity of inference.
Figure \ref{fig:model:mtl} shows a high-level overview of the \sys\ multi-task framework. 
%\luheng{TODO: Better high-level summary of how the model works}

%our model enumerates all possible mention and entity spans from the text, builds contextualized span representations from highway BiLSTM layers, and builds locally independent classifiers on top of the span pairs to decide whether there is a coreferent or relational link. 
%We also make use of the ELMo word embeddings \cite{peters2018deep}, we are able to achieve best results without any feature engineering. %, despite our small data situation.

\subsection{Problem Definition}
%\hanna{Luheng: can you swap the order of entity and relation in your definitions? I prefer to see entity come before than relations}
The input is a document represented as a sequence of words $D=\{w_1, \ldots, w_n\}$, from which we derive
$S=\{s_1,\ldots, s_N\}$, the set of all possible within-sentence word sequence spans (up to a reasonable length) in the document.
%, where each span is a sequence of words within a sentence.
The output contains three structures: 
the entity types $E$ for all spans $S$, 
the relations $R$ for all pair of spans $S\times S$, and the coreference links $C$ for all spans in $S$. 
The output structures are represented with a set of discrete random variables indexed by spans or pairs of spans.  
%In all three structures, we use $s_i$ to denote a span in the input document where
%$s_i=\{w_{\text{start}},\ldots, w_{\text{end}} \}\in\mathbf{S}$. We also assume spans do not cross sentence boundaries. $\mathbf{S}$ is the space of all spans in the documents.
%
%We use a shared model architecture for predicting all three kinds of structures: relations, entity types, and coreference clusters. 
Specifically, the output structures are defined as follows. 
%\luheng{TODO: explain more about candidate spans, unary and edge factors, etc.}
%
%\luheng{TODO: refer to the figure a bit.}

% \vspace{.1cm}
\noindent\textbf{Entity recognition}
is to predict the best entity type for every candidate span.
Let $L_{\text{E}}$ represent the set of all possible entity types including the null-type $\epsilon$. The output structure $E$ is a set of random variables indexed by spans: 
$e_i \in L_{\text{E}}$ for $i = 1 ,\ldots, N$. %, \forall 1\leq i\leq N$.
%$\{e_i \in L_{\text{E}} \mid 1\leq i\leq N\}$.
%\hanna{Can you edit the rest like entity recognition?}

% \vspace{.1cm}
\noindent\textbf{Relation extraction} 
is to predict  the best relation type given an ordered pair of spans $(s_i,s_j)$.
%For each pair of span $(s_i, s_j)$, the model predicts the best relation type from $s_i$ to $s_j$ \luheng{or reverse?}, or a special type indicating ``no relation''. 
Let $L_{\text{R}}$ be the set of all possible relation types including the null-type $\epsilon$. The output structure $R$ is a set of random variables indexed over pairs of spans $(i, j)$ that belong to the same sentence: 
%$\{r_{ij}\in L_{\text{R}} \mid 1\leq i,j\leq N, \text{sentence}(i)=\text{sentence}(j) \}$,
$r_{ij}\in L_{\text{R}}$ for $ i,j = 1, \ldots, N$.
%%MO: This is more confusing than helpful. You specify in words that they belong to the same sentence
%\text{sentence}(i)=\text{sentence}(j)$.
%Each random variable $r_{ij}$ can take value from the set of relation types $L_{\text{R}}$ as well as a ``no relation'' label $\epsilon$. 

% \vspace{.1cm}
\noindent\textbf{Coreference resolution} % \luheng{TODO}
is to predict the best antecedent (including a special null antecedent) given a span, which is the same mention-ranking model used in \newcite{Lee2017EndtoendNC}. 
The output structure $C$ is a set of random variables defined as:
$c_i \in \{1,\ldots, i-1, \epsilon \}$ for $i= 1,\ldots, N$.
%$\{ c_i \in \{1,\ldots, i-1, \epsilon \} \mid 1 \leq i \leq N \}$
%Same as \newcite{Lee2017EndtoendNC}, we use a mention-ranking model: For each span, the model predicts the best antecedent among all the preceding spans.\footnote{We limit the maximum number of antecedents following \newcite{lee2018higher}.} Therefore we have: $C=\{c_i\mid 1\leq i\leq m\}$, where $c_i \in \{1,\ldots, i-1, \epsilon \}$.

\subsection{Model Definition}
We formulate the multi-task learning setup as learning the conditional probability distribution $P(E,R,C|D)$. %over the three output structures $E$, $R$, and $C$.  
For efficient training and inference, we decompose $P(E,R,C|D)$ assuming spans are conditionally independent given $D$:
%this over the individual random variables:
\begin{align}
   %P(R \mid D) =& \prod_{1\leq i,j\leq N} P(r_{ij} \mid D)  \\
   %P(E \mid D) =& \prod_{1\leq i\leq N} P(e_i \mid D) \\
   %P(C \mid D) =& \prod_{1\leq i\leq N} P(c_i \mid D) 
    \hspace{-3cm}&P(E, R, C \mid D) = P(E, R, C, S \mid D) \\ 
    &=\prod_{i=1}^{N} 
  P(e_i \mid D) P(c_i \mid D) \nonumber 
   \prod_{j=1}^{N} P(r_{ij} \mid D),  \label{eq:decompose}
\end{align}
%\luheng{Shall we merge three structures into one probability? but then the weighted loss wouldn't work ...}
%where $r_{ij}$ $e_i$, and $c_i$ are random variables corresponding to the three tasks. 
where the conditional probabilities of each random variable are independently normalized:
\begin{align}
    & P(e_{i}=e\mid D) = \frac{
        \exp(\Phi_{\text{E}}(e, s_i))}{ \sum_{e'\in L_{\text{E}}} 
        \exp(\Phi_{\text{E}}(e', s_i))}  \\  \label{eq:normalizer}
    & P(r_{ij}=r\mid D) = \frac{
        \exp(\Phi_{\text{R}}(r, s_i, s_j))}{ \sum_{r'\in L_{\text{R}}} 
        \exp(\Phi_{\text{R}}(r', s_i, s_j))}  \notag \\
    & P(c_{i}=j\mid D) = \frac{
        \exp(\Phi_{\text{C}}(s_i, s_j))}{ \sum_{j' \in \{1,\ldots,i-1,\epsilon\}} \exp(\Phi_{\text{C}}(s_{i}, s_{j'}))}, \notag
\end{align}
where $\Phi_{\text{E}}$ denotes the unnormalized model score for an entity type $e$ and a span $s_i$, $\Phi_{\text{R}}$ denotes the score for a relation type $r$ and span pairs $s_i, s_j$, and $\Phi_{\text{C}}$ denotes the score for a binary coreference link between $s_i$ and $s_j$. 
These $\Phi$ scores are further decomposed into span and pairwise span scores computed from feed-forward networks, as will be explained in Section \ref{sec:model:scoring}.
 
For simplicity, we omit $D$ from the $\Phi$ functions and $S$ from the observation. 

\paragraph{Objective}
Given  a set of all documents $\mathcal{D}$, the model loss function is defined as a weighted sum of the negative log-likelihood loss of all three tasks:
\begin{align}
    %\mathcal{J}(R^*, E^*, C^*, D) =& 
 & -\sum_{(D, R^*, E^*, C^*) \in\mathcal{D}} \Big\{ 
 \lambda_{\text{E}}\log P (E^* \mid D)   \\
 & + \lambda_{\text{R}}\log P (R^* \mid D) + \lambda_{\text{C}}\log P (C^* \mid D) \Big\}\notag
\end{align}
where $E^*$, $R^*$, and $C^*$ are gold structures of the entity types, relations, and coreference, respectively. The task weights $\lambda_{\text{E}}$, $\lambda_{\text{R}}$, and $\lambda_{\text{C}}$ are introduced as hyper-parameters to control the importance of each task.

For entity recognition and relation extraction, $P(E^*\mid D)$ and $P(R^*\mid D)$ are computed with the definition in Equation \eqref{eq:decompose}.
For coreference resolution, we use the marginalized loss following \newcite{Lee2017EndtoendNC} since each mention can have multiple correct antecedents. Let $C^*_i$ be the set of all correct antecedents for span $i$, we have:
     $\log P (C^* \mid D)  = \sum_{i=1..N} \log \sum_{c\in C^*_i} P(c \mid D)$. %  \label{eq:coref_loss}
%\end{align}
%\hanna{why did we separate coref only here? how about the other two tasks} 
%\luheng{only coref uses marginalized loss}

\subsection{Scoring Architecture}\label{sec:model:scoring}
We use feedforward neural networks (FFNNs) over \emph{shared span representations} $\mathbf{g}$ to compute a set of span and pairwise span scores. 
For the span scores, $\phi_e(s_i)$ measures how likely a span $s_i$ has an entity type $e$, and $\phi_{\text{mr}}(s_i)$ and $\phi_{\text{mc}}(s_i)$ measure how likely a span $s_i$ is a mention in a relation or a coreference link, respectively. The pairwise scores $\phi_r(s_i, s_j)$ and $\phi_{\text{c}}(s_i, s_j)$ measure how likely two spans are associated in a relation $r$ or a coreference link, respectively. 
Let $\mathbf{g}_i$ be the fixed-length vector representation for span $s_i$.
For different tasks, the span scores $\phi_{\text{x}}(s_i)$ for $\text{x} \in \{e, \text{mc}, \text{mr}\}$ and pairwise span scores $\phi_{\text{y}}(s_i,s_j)$ for $\text{y} \in \{ r, \text{c}\}$ are computed as follows:
\begin{align*}
     \phi_{\text{x}} (s_i) =& \mathbf{w}_{\text{x}} \cdot \text{FFNN}_{\text{x}} (\mathbf{g}_i) \\
     \phi_{\text{y}} (s_i, s_j) =& \mathbf{w}_{\text{y}}\cdot \text{FFNN}_{\text{y}} ([
        \mathbf{g}_i, \mathbf{g}_j, \mathbf{g}_i \odot \mathbf{g}_j]),
\end{align*}
where $\odot$ is element-wise multiplication, and $\{ \mathbf{w}_{\text{x}}, \mathbf{w}_{\text{y}}\}$ are neural network parameters to be learned. 

We use these scores to compute the different $\Phi$:
\begin{eqnarray}
 \Phi_{\text{E}}(e, s_i) &= &\phi_{e}(s_i) \label{eq:factors} \\
 \Phi_{\text{R}}(r, s_i, s_j) &= & % \nonumber\\ 
\phi_{\text{mr}}(s_i) + \phi_{\text{mr}}(s_j)  + \phi_{r}(s_i, s_j) \notag \\
 \Phi_{\text{C}}(s_i, s_j) &=& \phi_{\text{mc}}(s_i) + %\notag\\ 
 \phi_{\text{mc}}(s_j) + \phi_{\text{c}}(s_i, s_j) \notag 
%\;\;e,r,j\neq \epsilon \notag
\end{eqnarray}

The scores in Equation \eqref{eq:factors} are defined for entity types, relations, and antecedents that are not the null-type $\epsilon$.
Scores involving the null label are set to a constant 0:
$ \Phi_{\text{E}}(\epsilon, s_i)=\Phi_{\text{R}}(\epsilon, s_i, s_j)=\Phi_{\text{C}}(s_i,\epsilon)=0$. 

We use the same span representations $\mathbf{g}$ from \cite{Lee2017EndtoendNC}
and share them across the three tasks.
We start by building bi-directional LSTMs \cite{hochreiter1997long} from word, character and ELMo \cite{peters2018deep} embeddings. 

For a span $s_i$, its vector representation $\mathbf{g}_i$ is constructed by concatenating $s_i$'s left and right end points from the BiLSTM outputs, an attention-based soft ``headword,'' and embedded span width features.
Hyperparameters and other implementation details will be described in Section \ref{sec:exp_setup}.

\subsection{Inference and Pruning}
Following previous work, we use beam pruning to reduce the number of pairwise span factors from $O(n^4)$ to $O(n^2)$ at both training and test time, where $n$ is the number of words in the document.
We define two separate beams: $B_{\text{C}}$ to prune spans for the coreference resolution task, and $B_{\text{R}}$ for relation extraction. The spans in the beams are sorted by their span scores $\phi_{\text{mc}}$ and $\phi_{\text{mr}}$ respectively, and the sizes of the beams are limited by $\lambda_{\text{C}} n$ and $\lambda_{\text{R}} n$. 
We also limit the maximum width of spans to a fixed number $W$, which further reduces the number of span factors to $O(n)$.

\section{Knowledge Graph Construction}
\label{sec:KG}

\begin{figure}[b]
\centering
\includegraphics%[width=8cm]
[width=\linewidth, keepaspectratio,trim={1.6cm 11cm 2.5cm 4cm},clip]{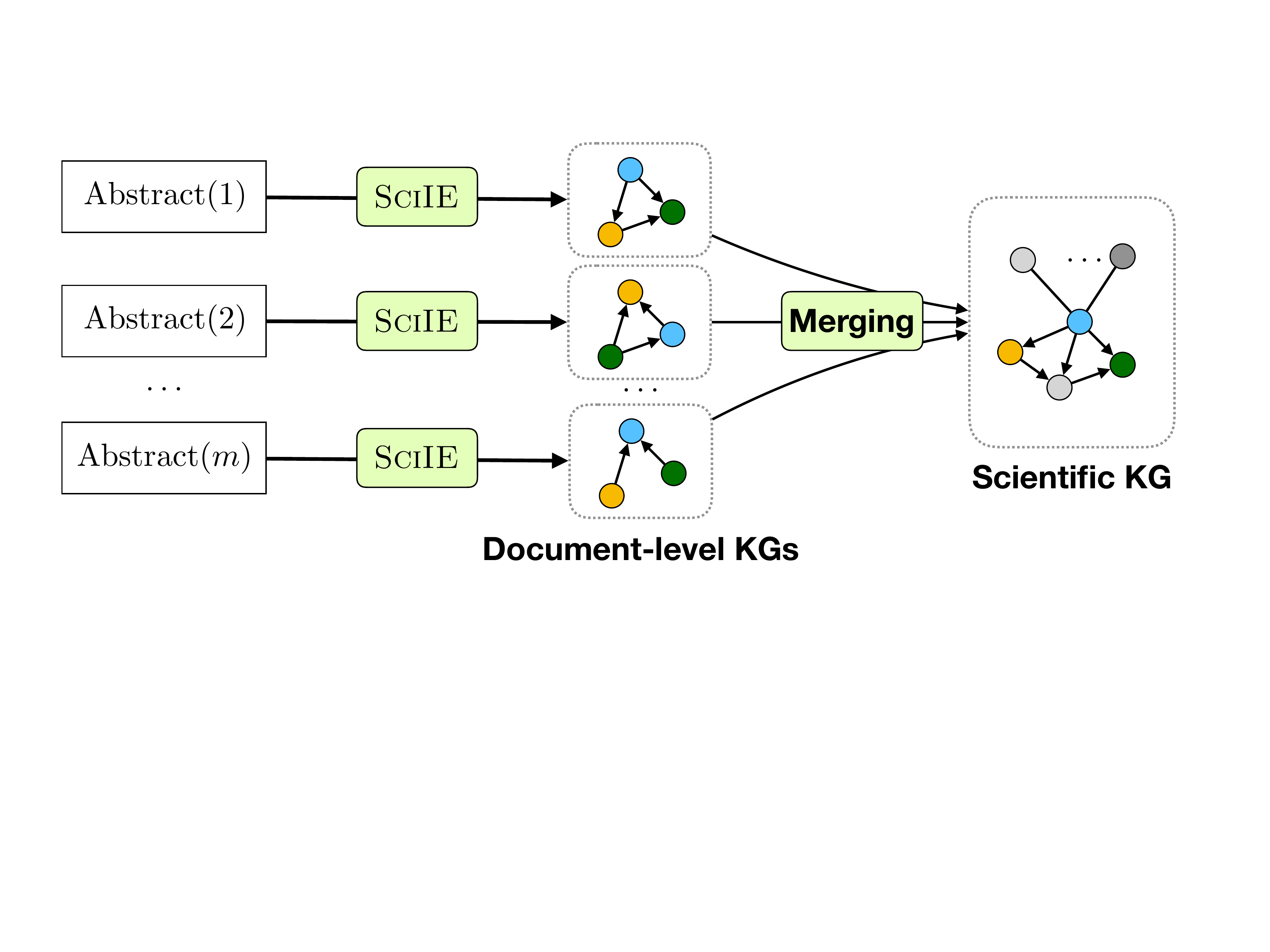}
% \vspace{-1em}
\caption{Knowledge graph construction process.}
\label{fig:KGconstruction}
% \vspace{-1em}
\end{figure}

\begin{figure}[t]
\centering
\includegraphics%[width=8cm]
[width=\linewidth, keepaspectratio,clip]{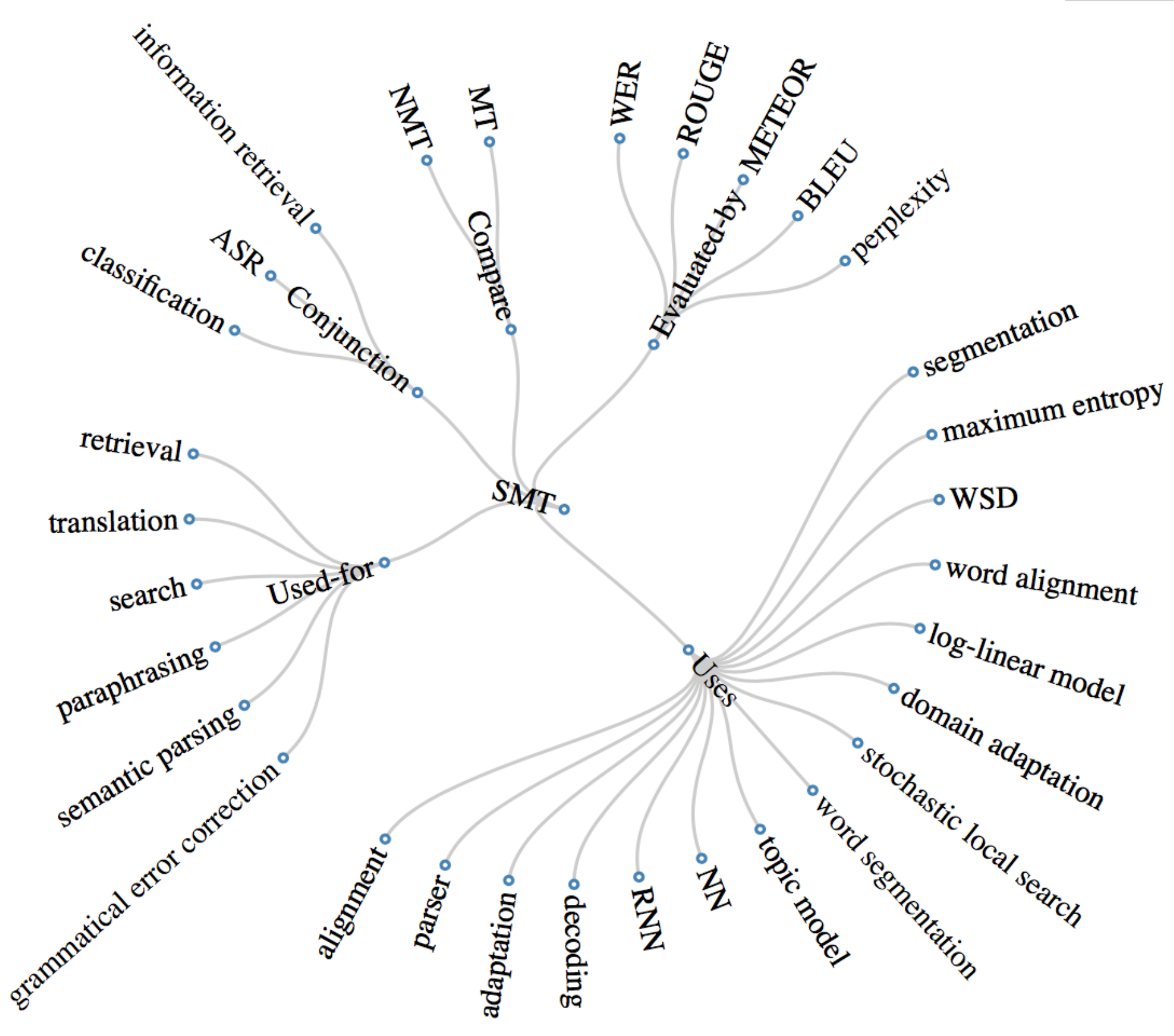}       
% \vspace{-1em}
\caption{A part of an automatically constructed scientific knowledge graph with the most frequent neighbors of the scientific term \textit{statistical machine translation (SMT)} on the graph. For simplicity we denote \textit{Used-for (Reverse)} as \textit{Uses}, \textit{Evaluated-for (Reverse)} as \textit{Evaluated-by}, and replace common terms with their acronyms. The original graph and more examples are given Figure \ref{fig:SMT} in Appendix \ref{sec:appendix}.}
\label{fig:KG}
% \vspace{-1em}
\end{figure}

We construct a scientific knowledge graph from a large corpus of  scientific articles. The corpus includes all abstracts (110k in total) from 12 AI conference proceedings from the Semantic Scholar Corpus. Nodes in the knowledge graph correspond to scientific entities. Edges correspond to scientific relations between pairs of entities. The edges are typed according to the relation types  defined in Section~\ref{sec:dataset}. Figure~\ref{fig:KG} shows a part of a knowledge graph created by our method. For example, \textit{Statistical Machine Translation (SMT)} and \textit{grammatical error correction} are nodes in the graph, and they are connected through a \textit{Used-for} relation type. 
%
%Our \sys\ model (described in Section~\ref{sec:model}) is developed to identify entities, relations, and coreference clusters within one document. 
In order to construct the knowledge graph for the whole corpus, we first apply the \sys\ model over  single documents and then integrate the entities and relations across multiple documents (Figure~\ref{fig:KGconstruction}).

\begin{figure}[t]
\hspace{-10pt}
\centering
\begin{tikzpicture}    
    \begin{axis}[
    	   width=.85\columnwidth,
	       height=.5\columnwidth,
	       font=\scriptsize,
            xbar,
            bar width=.18cm,
            y=.48cm,
            legend style={at={(0.6,1)},
                anchor=north,legend columns=-1},
            symbolic y coords={
                detection,
                object detection,
                face detection,
                human detection,
                pedestrian detection,
                action detection
            },
            ytick=data,
           % y tick label style={rotate=45, anchor=north east, inner sep=0mm},
            %y tick label style={rotate=30, anchor=north east, inner sep=0mm},
            %y tick label style={/pgf/number format/.cd,%
            %    set thousands separator={}, fixed, rotate=30},
            nodes near coords,
            nodes near coords style={/pgf/number format/.cd,%
                set thousands separator={}, fixed},%
            %nodes near coords={\pgfmathprintnumber\pgfplotspointmeta\%},
            %nodes near coords align={vertical,rotate=45},
            axis x line*=bottom,
            axis y line*=left,
            hide x axis,
            xmin=0,xmax=1400,
            xlabel near ticks,
            ylabel near ticks,
            xlabel={\# Entity Instances},
            legend image code/.code={
                \draw [#1] (0cm,-0.1cm) rectangle (0.3cm,0.08cm);
            }
        ]
        \addplot table[y=phrase,x=coref]{\detection};
        \addplot table[y=phrase,x=nocoref]{\detection};
        \legend{With Coref., Without Coref.}
        %\addplot table[x=interval,y=carD]{\nocoref};
        %addplot table[x=interval,y=carR]{\mydata};
        %\legend{Trips, Distance, Energy}
    \end{axis}
\end{tikzpicture}
% \vspace{-1em}
\caption{
Frequency of detected entities with and without coreferece resolution:
using coreference reduces the frequency of the generic phrase \textit{detection}  while significantly increasing the frequency of specific phrases.  Linking entities through coreference helps disambiguate phrases when generating the knowledge graph.
%\luheng{need to explan num. entities in text..}
}
\label{fig:KGEnt}
% \vspace{-1em}
\end{figure}
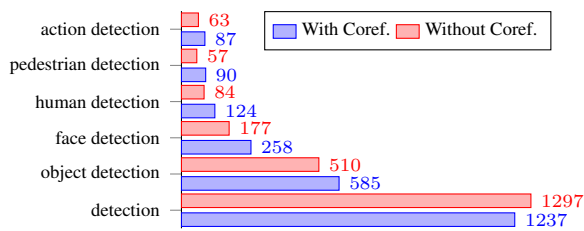

% \vspace{.1cm}
\paragraph{Extracting nodes (entities)} The \sys\ model  extracts entities, their relations, and coreference clusters within one document. 
%We normalize phrases in entities using some heuristics (described in implementation details) and through coreference links. 
Phrases are heuristically normalized (described in Section \ref{sec:exp_setup}) using entities and coreference links.
In particular, we link all entities that belong to the same coreference cluster to replace generic terms with any other non-generic term in the cluster. %We then list all the entities  entities across multiple documents with some string-matching heuristics (described in implementation details). 
Moreover, we replace all the entities in the cluster with the entity that has the longest string. Our qualitative analysis shows that there are fewer ambiguous phrases using coreference links (Figure~\ref{fig:KGEnt}). 
We calculate the frequency counts of all entities that appear in the whole corpus. We assign nodes in the knowledge graph by selecting the most frequent entities (with counts $>k$) in the corpus, and merge in any remaining entities for which a frequent entity is a substring.

\begin{figure}[t]
\hspace{-10pt}
\centering
\begin{tikzpicture}    
    \begin{axis}[
    	   width=.46\columnwidth,
	       height=.48\columnwidth,
	       font=\scriptsize,
	        ybar,
            bar width=.5cm,
            legend style={at={(0.8,1)},
                anchor=north,legend columns=-1},
            symbolic x coords={
                 Conjunction, Used for, Used for (Reverse)
            },
            xtick=data,
            enlarge x limits=0.3,
            x=0.7cm,
            x tick label style={rotate=18, anchor=north east, inner sep=0mm},
            nodes near coords,
            nodes near coords align={vertical},
            axis x line*=bottom,
            axis y line*=left,
            %hide y axis,
           % xmin=-1,ymax=4,
            ymin=0,ymax=90,
            ylabel near ticks,
            ylabel={\# Relation Triples},
            legend image code/.code={
                \draw [#1] (0cm,-0.1cm) rectangle (0.1cm,0.2cm);
            }
        ]
        \addplot table[x=relation,y=freq]{\mtasr};
        \legend{MT-ASR}
        %\legend{Machine Translation vs. Automatic Speech Recognition}
    \end{axis}
\end{tikzpicture}
\hspace{-5pt}
\begin{tikzpicture}    
    \begin{axis}[
    	   width=.54\columnwidth,
	       height=.48\columnwidth,
	       font=\scriptsize,
            ybar,
            bar width=.5cm,
            legend style={at={(0.7,1)},
                anchor=north,legend columns=-1},
            symbolic x coords={
                 Hyponym of, Conjunction, Used for, Used for (Reverse)
            },
            xtick=data,
            x tick label style={rotate=18, anchor=north east, inner sep=0mm},
            nodes near coords,
            %nodes near coords={\pgfmathprintnumber\pgfplotspointmeta\%},
            %nodes near coords align={vertical},
            enlarge x limits=0.3,
            x=0.7cm,
            axis x line*=bottom,
            axis y line*=left,
            %hide y axis,
            ymin=0,ymax=30,
            legend image code/.code={
                \draw [#1] (0cm,-0.1cm) rectangle (0.1cm,0.2cm);
            }
        ]
        \addplot table[x=relation,y=freq]{\crfgm};
        \legend{CRF-GM}
    \end{axis}
\end{tikzpicture}
% \vspace{-1em}
\caption{Frequency of relation types between pairs of entities: ({\it left}) automatic speech recognition (ASR) and machine translation (MT), ({\it right}) conditional random field (CRF) and graphical model (GM).   We use the most frequent relation between pairs of entities in the knowledge graph.  %; Similalrly, there are multiple relations The most frequent relation between machine translation (MT) and automatic speech recognition (ASR) is \textit{Conjunction}, but sometimes MT may also be {\it used for ASR or used by ASR. Similarly, conditional random field (CRF) can be a hyponym of graphical model (GM) or a conjunction with GM. %We use the number of relation triple occurrence as our confidence score and choose the most confident relation type to build the knowledge graph.
}
\label{fig:KGRel}
% \vspace{-1em}
\end{figure}
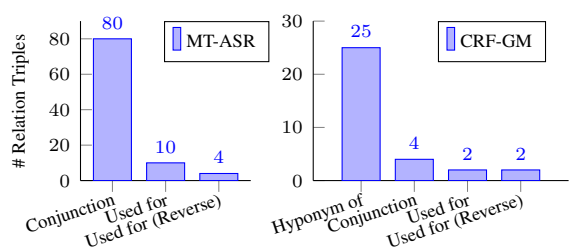

% \vspace{.1cm}
\paragraph{Assigning edges (relations)} 
A pair of entities may appear in different contexts, resulting in different relation types between those entities (Figure~\ref{fig:KGRel}).
For every pair of entities in the graph,  we calculate the frequency of different relation types across the whole corpus.%, and assign edge by selecting the most frequent type.
We assign edges between entities by selecting the most frequent relation type.

\pgfplotstableread[row sep=\\,col sep=&]{
    phrase               & coref & nocoref \\
    detection            & 1237 & 1297 \\
    object detection     & 585  & 510  \\
    face detection       & 258  & 177  \\
    human detection      & 124  & 84   \\
    pedestrian detection & 90   & 57   \\
    action detection     & 87   & 63   \\
    }\detection

\pgfplotstableread[row sep=\\,col sep=&]{
    relation             & freq \\
    Conjunction    &  80  \\
    Used for       &  10  \\
    Used for (Reverse)    &   4  \\
    }\mtasr
    
\pgfplotstableread[row sep=\\,col sep=&]{
    relation             & freq \\
    Hyponym of     &   25 \\
    Conjunction    &    4 \\
    Used for       &    2 \\
    Used for (Reverse)   &   2 \\
    }\crfgm

\section{Experimental Setup}\label{sec:exp_setup}
We evaluate our unified framework \sys\ on \dataname\ and \semevalEnt. The knowledge graph for scientific community analysis is built using the Semantic Scholar Corpus (110k abstracts in total).

\subsection{Baselines}

We compare our model with the following baselines on \dataname dataset:
 \begin{itemize}
 \item \textbf{LSTM+CRF} The state-of-the-art NER system~\cite{lample2016neural}, which applies CRF on top of LSTM for named entity tagging,  the approach has also been  used in scientific term extraction~\cite{luan2017scienceie}.
  \item \textbf{LSTM+CRF+ELMo} LSTM+CRF with \elmo\ as an additional input feature.
\item \textbf{\mb} State-of-the-art joint entity and relation extraction system~\cite{miwa2016end} that has also been  used in scientific literature~\cite{peters2017semi,scienceIE}.  This system 
%%MO: commented out, redundant
%(denoted as \mb) 
uses syntactic features such as part-of-speech tagging and dependency parsing. 
\item \textbf{\mb(Pipeline)} Pipeline setting of \mb. Extract entities first and use entity results as input to relation extraction task.
\item \textbf{E2E Rel+ELMo} \mb\ with \elmo\ as an additional input feature.
\item \textbf{\coref} State-of-the-art coreference system \newcite{Lee2017EndtoendNC} combined with \elmo. Our system \sys\ extends E2E Coref with multi-task learning.

\end{itemize}

In the SemEval task, we compare our model \sys\ with the best reported system in the SemEval leaderboard~\cite{peters2017semi}, which extends \mb\ with several in-domain features such as 
gazetteers extracted from existing knowledge bases and model ensembles.
We also compare with the state of the art on keyphrase extraction~\cite{luan2017scienceie}, which applies semi-supervised methods to a neural tagging model.\footnote{We compare with the inductive setting results.}

%  \vspace{-.2cm}
\subsection{Implementation details}  
Our system extends the implementation and hyper-parameters from \newcite{Lee2017EndtoendNC} with the following adjustments. We use a 1 layer BiLSTM with 200-dimensional hidden layers. All the FFNNs have 2 hidden layers of 150 dimensions each.  We use 0.4 variational dropout  \cite{gal:2016} for the LSTMs, 0.4 dropout for the FFNNs, and 0.5 dropout for the input embeddings. 
We model spans up to 8 words. For beam pruning, we use $\lambda_{\text{C}}=0.3$ for coreference resolution and $\lambda_{\text{R}}=0.4$ for relation extraction.
For constructing the knowledge graph, we use the following heuristics to normalize the entity phrases. We replace all acronyms with their corresponding full name and normalize all the plural terms with their singular counterparts.
\begin{table}[t]
\newcolumntype{Y}{>{\centering\arraybackslash}X}
\newcommand{\colindent}{\;}
\setlength{\tabcolsep}{.25em}
\footnotesize
\centering

\begin{subtable}{\linewidth}
\begin{tabularx}{\linewidth}{l c *{6}{Y}}
\toprule
& \multicolumn{3}{c}{Dev} & \multicolumn{3}{c}{Test} \\
\cmidrule(lr){2-4} \cmidrule(lr){5-7} 
Model & P & R & F1 & P & R & F1 \\    
\midrule
LSTM+CRF & 67.2   & 65.8   & 66.5  & 62.9   & 61.1   & 62.0   \\
LSTM+CRF+ELMo & 68.1   & 66.3   & 67.2  & 63.8   & 63.2   & 63.5   \\
\mb({\pipeline}) & 66.7   & 65.9   & 66.3  & 60.8   & 61.2   & 61.0   \\
\mb     & 64.3   & 68.6   & 66.4  & 60.6   & 61.9   & 61.2   \\
\mb+\elmo & 67.5   & 66.3   & 66.9  & 63.5   & 63.9   & 63.7  \\
%\cmidrule(lr){1-7}
\sys\ & 70.0   & 66.3  & \textbf{68.1}  & 67.2   &  61.5  & \textbf{64.2}  \\
\bottomrule
\end{tabularx}
\caption{Entity recognition.}\label{tab:ner}
\end{subtable}

\begin{subtable}{\linewidth}
\begin{tabularx}{\columnwidth}{l c *{6}{Y}}
\toprule
& \multicolumn{3}{c}{Dev} & \multicolumn{3}{c}{Test} \\
\cmidrule(lr){2-4} \cmidrule(lr){5-7} 
Model & P & R & F1 & P & R & F1 \\    
\midrule
\mb({\pipeline})  & 34.2   & 33.7   & 33.9  & 37.8   & 34.2   & 35.9  \\

\mb   & 37.3   & 33.5   & 35.3  & 37.1   & 32.2   & 34.1  \\
\mb+\elmo & 38.5   & 36.4   & 37.4  & 38.4   & 34.9   & 36.6  \\
%\cmidrule(lr)
\sys\ & 45.4   & 34.9  & \textbf{39.5}  & 47.6   &  33.5  & \textbf{39.3}  \\
\bottomrule
 \end{tabularx}
 \caption{Relation extraction.}\label{tab:relation}
 \end{subtable}
 
 \begin{subtable}{\linewidth}
 \begin{tabularx}{\linewidth}{l c *{6}{Y}}
\toprule
& \multicolumn{3}{c}{Dev} & \multicolumn{3}{c}{Test} \\
\cmidrule(lr){2-4} \cmidrule(lr){5-7} 
     Model & P & R & F1 & P & R & F1 \\    
    \midrule
{\coref}     & 59.4   & 52.0   & 55.4  & 60.9   & 37.3   & 46.2    \\
\sys & 61.5   & 54.8   & \textbf{58.0}  & 52.0 & 44.9 & \textbf{48.2} \\
\bottomrule
\end{tabularx}
\caption{Coreference resolution. 
}\label{tab:coref}
\end{subtable}
% \vspace{-1em}
\caption{
Comparison with previous systems on the development and test set for our three tasks.
For coreference resolution, we report the average P/R/F1 of MUC, $\text{B}^3$, and $\text{CEAF}_{\phi_4}$ scores. %\newcite{Pradhan2012CoNLL2012ST}.
}
\label{tab:alltasks}
% \vspace{-1em}
\end{table}

\section{Experimental Results}\label{sec:exp_result}

We evaluate \sys\ on \dataname\ and  \semevalEnt\ datasets. We  provide qualitative results and human evaluation of the constructed knowledge graph. 

\subsection{IE Results}
% \paragraph{Entity and Relation Extraction}
\paragraph{Results on SciERC} Table~\ref{tab:alltasks} compares the result of our model with baselines on the three tasks: entity recognition (Table~\ref{tab:ner}), relation extraction (Table~\ref{tab:relation}), and coreference resolution (Table~\ref{tab:coref}).  As evidenced by the table, our unified multi-task setup \sys\ outperforms all the baselines.
For entity recognition, our model achieves 1.3\% and 2.4\% relative improvement over LSTM+CRF with and without \elmo, respectively. 
%. Our model achieves 
Moreover, it achieves 1.8\% and 2.7\% relative improvement over \mb\ with and without \elmo, respectively. For relation extraction, we observe more significant improvement with  13.1\% relative improvement over \mb\ and 7.4\% improvement over \mb\ with \elmo. For coreference resolution, \sys\ outperforms \coref\ with 4.5\% relative improvement.  We still observe a large gap between human-level performance and a machine learning system. We invite the community to address this challenging task.

\begin{table}[t]
\newcolumntype{Y}{>{\centering\arraybackslash}X}
\setlength{\tabcolsep}{.25em}
\footnotesize
\begin{tabularx}{\linewidth}{l c *{3}{Y}}
      \toprule
     %Task & \multicolumn{1}{c}{Single Task} & \multicolumn{1}{c}{w/ ner} & \multicolumn{1}{c}{w/ relation} & \multicolumn{1}{c}{w/ coref} & \multicolumn{1}{c}{w/o elmo}\\    
   Task &  Entity Rec. & Relation  & Coref.  \\
    %\cmidrule(lr){2-4} \cmidrule(lr){5-7} \cmidrule(lr){8-10} \cmidrule{11-13} \cmidrule{14-16}
    %Metric & P & R & F1 & P & R & F1 &P & R & F1 &P & R & F1 & P & R & F1 \\  
    \midrule  
%Best Multitask   & 68.1 & 40.6 & 58.6 \\
%$-$\texttt{ELMo} & 65.2 & 34.5 & 52.6 \\
%\cmidrule(lr){1-4}
Multi Task (\sys) & 68.1 & 39.5 & 58.0 \\
    \midrule  
Single Task & 65.7 & 37.9 & 55.3 \\
$+$Entity Rec.      & -    & 38.9 & 57.1 \\
$+$Relation & 66.8 & -    & 57.6 \\
$+$Coreference    & 67.5 & 39.5 & -    \\
\bottomrule

\end{tabularx}
% \vspace{-1em}
\caption{Ablation study for multitask learning on \dataname\ development set. Each column shows results for the target task. %with ablations on different multitask setups.
}
  \label{tab:ablation}
%   \vspace{-1em}
\end{table}

% \begin{table*}[t]
% \newcolumntype{Y}{>{\centering\arraybackslash}X}
% \setlength{\tabcolsep}{.25em}
% \footnotesize
% \begin{tabularx}{\textwidth}{l c *{15}{Y}}
%   %\begin{tabular}{l|lll|lll|lll|lll|lll}
%       \toprule
%      Task & \multicolumn{3}{c}{Single Task} & \multicolumn{3}{c}{w/ ner} & \multicolumn{3}{c}{w/ relation} & \multicolumn{3}{c}{w/ coref} & \multicolumn{3}{c}{w/o elmo}\\    
%     \cmidrule(lr){2-4} \cmidrule(lr){5-7} \cmidrule(lr){8-10} \cmidrule{11-13} \cmidrule{14-16}
%     Metric & P & R & F1 & P & R & F1 &P & R & F1 &P & R & F1 & P & R & F1 \\  
%     \midrule  
% ner & 66.1   & 65.2   & 65.7 & - & - & - & 67.8 & 65.8 & 66.8 & 69.2 & 65.8 & 67.5 & 66.3 & 64.1 & 65.2\\
% relation & 44.2 & 33.2 & 37.9 & 42.9 & 35.6 & 38.9 &- & - & -  & 45.5 & 36.6 & 40.6 & 39.7 & 30.5 & 34.5\\
% coref & 66.0 & 47.6 & 55.3 & 66.2 & 50.3 & 57.1 & 61.6 & 54.8 & 58.0 & - & - & - & 65.7 &43.8 &52.6\\
% \bottomrule
% \end{tabularx}
  
% \caption{Ablation study for multitask learning on the XXX development set. Each row shows results for the target task, with ablations on different multitask setups. We also show ablations by removing the ELMo embeddings.
% \luheng{say more, also change task names} \hanna{The table is hard to read and compare; I prefer to see the tasks in the top and the ablations as rows}}
%   \label{tab:ablation}
% \end{table*}

\begin{table*}[t]
\newcolumntype{Y}{>{\centering\arraybackslash}X}
\setlength{\tabcolsep}{.25em}
\footnotesize
\begin{tabularx}{\textwidth}{l c *{12}{Y}}
\toprule
    & \multicolumn{3}{c}{Span Indentification} & \multicolumn{3}{c}{Keyphrase Extraction} & \multicolumn{3}{c}{Relation Extraction} & \multicolumn{3}{c}{Overall}\\
    \cmidrule(lr){2-4} \cmidrule(lr){5-7} \cmidrule(lr){8-10} \cmidrule{11-13}
     Model & P & R & F1 & P & R & F1 & P & R & F1 & P & R & F1\\    
    \midrule
(Luan 2017)   & -   & -  & 56.9  & - & - & 45.3 & - &- &- & - &- &-\\
Best SemEval & 55   & 54   & 55  & 44 & 43 & 44 & 36 & 23 & \textbf{28} & 44 & 41 & 43\\
\sys\ & 62.2   & 55.4  & \textbf{58.6}  & 48.5   &  43.8  & \textbf{46.0} & 40.4 & 21.2 & 27.8 & 48.1 & 41.8 & \textbf{44.7} \\
\bottomrule
\end{tabularx}
% \vspace{-1em}
\caption{Results for scientific keyphrase extraction and extraction on SemEval 2017 Task 10, comparing with previous best systems.}
  \label{tab:semeval}
%   \vspace{-1em}
\end{table*}

\paragraph{Ablations}
We evaluate the effect of multi-task learning in each of the three tasks defined in our dataset. 
Table~\ref{tab:ablation} reports the results for individual tasks when additional tasks are included in the learning objective function.
We observe that performance improves with each added task in the objective.
For example, Entity recognition (65.7) benefits from both coreference resolution (67.5) and relation extraction (66.8). Relation extraction (37.9)  significantly benefits when multi-tasked with coreference resolution  (7.1\% relative improvement). Coreference resolution benefits when multi-tasked with  relation extraction, with 4.9\% relative improvement.

\paragraph{Results on \semevalEnt}
Table~\ref{tab:semeval} compares the results of our model with the state of the art on the \semevalEnt\ dataset for tasks of span identification, keyphrase extraction and relation extraction as well as the overall score.  Span identification aims at  identifying spans of entities. Keyphrase classification and relation extraction has the same setting with the entity and relation extraction in \dataname. Our model outperforms all the previous models that use hand-designed features.
We observe more significant improvement in span identification than keyphrase classification. This confirms the benefit of our model in enumerating spans (rather than BIO tagging in state-of-the-art systems).
Moreover, we  have competitive results compared to the previous state of the art in relation extraction. We observe less gain compared to the \dataname\ dataset mainly because there are no coference links, and  the  relation types are not comprehensive.
% \section{Analysis}
% \begin{figure}[t]
% \centering
% \includegraphics[width=7.5cm]{./figures/detection.pdf}           
% % \caption{Propagation through coreference links can reduce the disambiguity when generating the knowledge graph. The number of more general phrase \textit{parsing} reduced while the more specific terms \textit{semantic parsing}, \textit{dependency parsing} and \textit{syntactic parsing} grow significantly}
% % \label{fig:KGterm}
% \caption{Propagation through coreference links can reduce the disambiguity when generating the knowledge graph. The number of more general phrase \textit{detection} reduced while the more specific terms \textit{object detection}, \textit{face detection}, etc. grow significantly
% }
% \label{fig:KGterm}
% \end{figure}
\subsection{Knowledge Graph Analysis}
 We provide qualitative analysis and human evaluations on the constructed knowledge graph. 
\pgfplotstableread[row sep=\\,col sep=&]{
year & mt & lm & pos \\
1996 & 0.04 & 0.00 & 0.00 \\
1997 & 0.02 & 0.00 & 0.00 \\
1998 & 0.00 & 0.00 & 0.00 \\
1999 & 0.02 & 0.00 & 0.00 \\
2000 & 0.03 & 0.02 & 0.00 \\
2001 & 0.02 & 0.00 & 0.00 \\
2002 & 0.01 & 0.00 & 0.00 \\
2003 & 0.02 & 0.02 & 0.00 \\
2004 & 0.04 & 0.00 & 0.00 \\
2005 & 0.02 & 0.00 & 0.04 \\
2006 & 0.03 & 0.00 & 0.00 \\
2007 & 0.01 & 0.00 & 0.00 \\
2008 & 0.02 & 0.00 & 0.00 \\
2009 & 0.01 & 0.00 & 0.02 \\
2010 & 0.06 & 0.01 & 0.00 \\
2011 & 0.10 & 0.01 & 0.02 \\
2012 & 0.12 & 0.01 & 0.02 \\
2013 & 0.26 & 0.03 & 0.09 \\
2014 & 0.27 & 0.07 & 0.08 \\
2015 & 0.46 & 0.18 & 0.19 \\
2016 & 0.58 & 0.53 & 0.28 \\
}\nlp

\pgfplotstableread[row sep=\\,col sep=&]{
year & sr & ss & spr \\
1996 & 0.14 & 0.00 & 0.00 \\
1997 & 0.14 & 0.08 & 0.07 \\
1998 & 0.08 & 0.17 & 0.00 \\
1999 & 0.06 & 0.00 & 0.00 \\
2000 & 0.04 & 0.04 & 0.00 \\
2001 & 0.04 & 0.01 & 0.14 \\
2002 & 0.03 & 0.00 & 0.00 \\
2003 & 0.02 & 0.02 & 0.02 \\
2004 & 0.01 & 0.04 & 0.07 \\
2005 & 0.02 & 0.02 & 0.00 \\
2006 & 0.03 & 0.00 & 0.00 \\
2007 & 0.03 & 0.02 & 0.02 \\
2008 & 0.03 & 0.02 & 0.03 \\
2009 & 0.01 & 0.01 & 0.00 \\
2010 & 0.05 & 0.00 & 0.05 \\
2011 & 0.05 & 0.01 & 0.01 \\
2012 & 0.08 & 0.00 & 0.00 \\
2013 & 0.21 & 0.03 & 0.02 \\
2014 & 0.32 & 0.11 & 0.10 \\
2015 & 0.43 & 0.23 & 0.20 \\
2016 & 0.51 & 0.61 & 0.50 \\
}\speech

\pgfplotstableread[row sep=\\,col sep=&]{
year & or & od & is \\
1996 & 0.00 & 0.00 & 0.00 \\
1997 & 0.00 & 0.00 & 0.00 \\
1998 & 0.00 & 0.00 & 0.00 \\
1999 & 0.00 & 0.00 & 0.00 \\
2000 & 0.00 & 0.00 & 0.00 \\
2001 & 0.00 & 0.00 & 0.00 \\
2002 & 0.00 & 0.00 & 0.00 \\
2003 & 0.00 & 0.00 & 0.00 \\
2004 & 0.00 & 0.00 & 0.08 \\
2005 & 0.00 & 0.00 & 0.00 \\
2006 & 0.00 & 0.00 & 0.00 \\
2007 & 0.00 & 0.00 & 0.00 \\
2008 & 0.03 & 0.00 & 0.00 \\
2009 & 0.02 & 0.00 & 0.02 \\
2010 & 0.00 & 0.00 & 0.00 \\
2011 & 0.03 & 0.00 & 0.00 \\
2012 & 0.06 & 0.00 & 0.00 \\
2013 & 0.06 & 0.04 & 0.00 \\
2014 & 0.18 & 0.16 & 0.00 \\
2015 & 0.35 & 0.27 & 0.29 \\
2016 & 0.50 & 0.37 & 0.37 \\
}\cv

\begin{figure}[t]
\centering
\begin{tikzpicture}    
\hspace{-5pt}
    \begin{axis}[
    	   width=\columnwidth,
	       height=.45\columnwidth,
	       font=\scriptsize,
            legend style={at={(0.05,1.05)}, anchor=north west}, %,legend columns=-1},
            nodes near coords align={vertical},
            %axis x line*=bottom,
            %axis y line*=left,
            %hide x axis,
            ymin=-0.05,ymax=0.60,
            %ylabel near ticks,
            %ylabel={Freq},
            mark options={mark size=1},
        ]
        \addplot[smooth,mark=square,blue,thick]  table[x=year,y=mt]{\nlp};
        \addplot[smooth,mark=o,red,thick]  table[x=year,y=lm]{\nlp};
        \addplot[smooth,mark=diamond,g-green,thick]  table[x=year,y=pos]{\nlp};
        \legend{Language Modeling, Machine Translation , POS Tagging}
    \end{axis}
\end{tikzpicture}
\begin{tikzpicture}    
% \vspace{-25pt}
\hspace{-5pt}
    \begin{axis}[
    	   width=\columnwidth,
	       height=.45\columnwidth,
	       font=\scriptsize,
            legend style={at={(0.05,1.05)}, anchor=north west}, %,legend columns=-1},
            nodes near coords align={vertical},
            %axis x line*=top,
            %axis y line*=left,
            %hide x axis,
            ymin=-0.05,ymax=0.65,
            %ylabel near ticks,
            %ylabel={Freq}
            mark options={mark size=1},
        ]
        \addplot[smooth,mark=square,blue,thick] table[x=year,y=sr]{\speech};
        \addplot[smooth,mark=o,red,thick]  table[x=year,y=ss] {\speech};
        \addplot[smooth,mark=diamond,g-green,thick] table[x=year,y=spr] {\speech};
        \legend{Speech Recognition, Speech Synthesis, Speaker Recognition}
    \end{axis}
\end{tikzpicture}
\begin{tikzpicture}    
% \vspace{-25pt}
\hspace{-5pt}
    \begin{axis}[
    	   width=\columnwidth,
	       height=.45\columnwidth,
	       font=\scriptsize,
            legend style={at={(0.05,1.05)}, anchor=north west}, %,legend columns=-1},
            nodes near coords align={vertical},
            %axis x line*=bottom,
            %axis y line*=left,
            ymin=-0.05,ymax=0.52,
            %ylabel near ticks,
            %ylabel={Freq},
            mark options={mark size=1},
            x tick label style={/pgf/number format/.cd,%
                set thousands separator={},
                fixed}%
        ]
        \addplot[smooth,mark=square,blue,thick] table[x=year,y=or]{\cv};
        \addplot[smooth,mark=o,red,thick] table[x=year,y=od]{\cv};
        \addplot[smooth,mark=diamond,g-green,thick] table[x=year,y=is]{\cv};
        \legend{Object Recognition, Object Detection, Image Segmentation}
    \end{axis}
\end{tikzpicture}
% \vspace{-1em}
\caption{
Historical trend for top applications of the keyphrase \emph{neural network} in NLP, speech, and CV conference papers we collected. y-axis indicates the ratio of papers that use \emph{neural network} in the task to the number of papers that is about the task.
}
\label{fig:KGtrend}
% \vspace{-1em}
\end{figure}
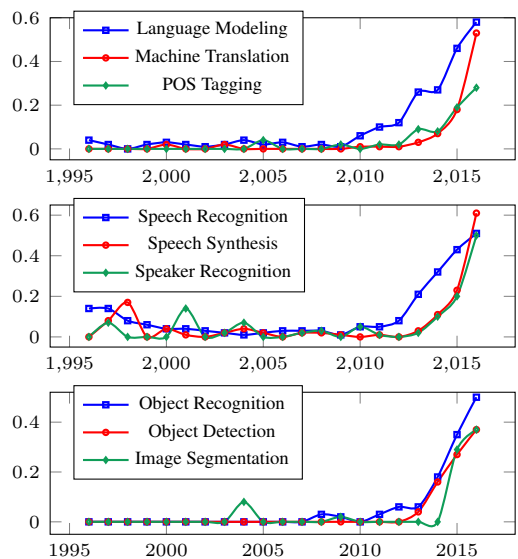 
\begin{figure}[t]
\begin{tikzpicture}
\begin{axis}[
    	width=1\columnwidth,
	    height=0.5\columnwidth,
	    legend style={at={(1, 1)},anchor=north east,font=\scriptsize},
	    mark options={mark size=2},
		font=\scriptsize,
		xlabel near ticks,
		ylabel near ticks,
	    xmin=0,xmax=102,
   		ymin=83.5, ymax=92,
   	 	ymajorgrids=true,
    	xmajorgrids=true,
    	xlabel=Pseudo-recall \%,
        ylabel=Precision \%,
    	ylabel style={yshift=-1.5ex,}]

    \addplot[smooth,mark=o,blue,thick] plot coordinates {
    (11.18,89.96)
(14.41,89.63)
(14.41,89.63)
(19.14,90.20)
(19.14,90.20)
(24.95,89.55)
(24.95,89.55)
(34.41,89.89)
(34.41,89.89)
(55.05,88.07)
(55.05,88.07)
(96.56,84.93)
    };
    \addlegendentry{With Coref.}
    
    \addplot[smooth,mark=square,red,thick] plot coordinates {
      (9.03,89.36)
(11.40,89.83)
(11.40,89.83)
(15.70,90.12)
(15.70,90.12)
(20.00,88.57)
(20.00,88.57)
(28.82,87.01)
(28.82,87.01)
(48.82,85.98)
(48.82,85.98)
(89.68,84.07)
    };
    \addlegendentry{Without Coref.}
    
    \end{axis}
    
\end{tikzpicture}
% \vspace{-1em}
\caption{Precision/pseudo-recall curves for human evaluation by varying cut-off thresholds. The  AUC is 0.751 with coreference, and 0.695 without.
\label{fig:humaneval}}
% \vspace{-1em}
\end{figure}
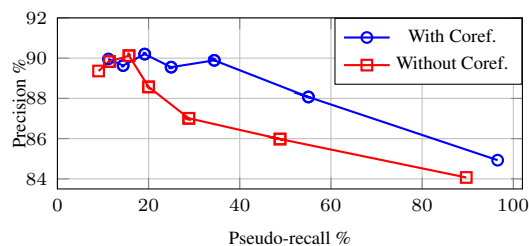

% \vspace{-.1cm}
\paragraph{Scientific trend analysis}
Figure~\ref{fig:KGtrend} shows the historical trend analysis (from 1996 to 2016) of the most popular applications of the phrase \textit{neural network}, 
%The popular applications of  the  \textit{neural network}  method are 
selected according to the statistics of the extracted relation triples with the {`Used-for'} relation type from speech, computer vision, and NLP conference papers.  We  observe that, before 2000, \textit{neural network} has been applied to a greater percentage of speech applications compared to the NLP and computer vision papers. In NLP, neural networks first gain popularity in language modeling and then extend to other tasks such as POS Tagging and Machine Translation. In computer vision, the application of neural networks gains popularity in \textit{object recognition} earlier (around 2010) than the other two more complex tasks of \textit{object detection} and \textit{image segmentation} (hardest and also the latest). 

% \vspace{-.1cm}
\paragraph{Knowledge Graph Evaluation}
Figure~\ref{fig:humaneval} shows the human evaluation of the constructed knowledge graph,  comparing the quality of automatically generated knowledge graphs with and without the coreference links. 
We randomly select 10 frequent scientific entities and extract all the relation triples  that include one of the selected entities leading to  1.5k relation triples from both systems.  %target entity involved. We filter out entities that has less than 3 co-occurrence with the target entity and choose the most confident relation as edge type (as described in Section~\ref{sec:KG}).  \hanna{I didn't understand how you selected the neighbors?} \ylcomment{Does this answer?}
We ask four domain experts to annotate each of these extracted relations to define ground truth labels.
Each domain expert is assigned 2 or 3  entities and all of the corresponding relations. 
%
%We define ground truth labels as the human annotations and compute precision and recall. 
Figure~\ref{fig:humaneval} shows precision/recall curves for both systems. Since it is not feasible to compute the actual recall of the systems, we compute the pseudo-recall \cite{Zhang2015ExploitingPN} based on the output of both systems.  
%\hanna{cite to an IE paper, that uses  pseudo recall; look at Dan Weld's papers for relation extraction}
%
We  observe that the knowledge graph curve with coreference linking is  mostly above the curve without coreference linking.  The precision of both systems is high (above 84\% for both systems), but the system with coreference links has significantly higher recall.

\section{Conclusion}
In this paper, we create a new dataset and develop a multi-task model for identifying entities, relations, and coreference clusters in scientific articles. 
By sharing span representations and leveraging cross-sentence information, our multi-task setup effectively improves performance across all tasks. 
Moreover, we show that our multi-task model is better at predicting span boundaries and outperforms previous state-of-the-art scientific IE systems on entity and relation extraction, without using any hand-engineered features or pipeline processing.
%and could be extended to other scientific domains with sufficient training data.
%
Using our model, we are able to automatically organize the extracted information from a large collection of scientific articles into a knowledge graph.
Our analysis shows the importance of coreference links in making a dense, useful graph. %It additionally shows interesting observations about the scientific trend. 

We still observe a large gap between the performance of our model and human performance, confirming the challenges of scientific IE.
Future work includes improving the performance using semi-supervised techniques and providing in-domain features. We also plan to extend our multi-task framework to information extraction tasks in other domains. 
\subsection*{Acknowledgments}
This research was supported by the Office of Naval Research under the MURI grant N00014-18-1-2670, NSF (IIS 1616112, III 1703166), Allen Distinguished Investigator Award, and gifts from Allen Institute for AI, Google, Amazon, and Bloomberg. We are grateful to Waleed Ammar and AI2 for sharing the Semantic Scholar Corpus. We also thank the anonymous reviewers, UW-NLP group and Shoou-I Yu for their helpful comments.

% This research was supported by the NSF (IIS 1616112), Allen Institute for AI (66-9175), Allen Distinguished Investigator Award, and gifts from Google, Samsung, and Bloomberg. We thank the anonymous reviewers for their helpful comments.
\bibliography{references}
\bibliographystyle{emnlp_natbib}
% \begin{table}[t]
%   \centering
%  {\small
%   \begin{tabular}{l|llll}
%     \toprule
%      Model & Overlap & Correct & False hit & Miss \\    
%     \midrule
%      SL & 568 & 1064 & 668 & 395\\
%      SSL & 533 & 1193 & 747 & 318\\
%     \bottomrule
%   \end{tabular}}
%   \caption{Error analysis on keyword identification task}
%   \label{tab:span}
% \end{table}
\newpage
\newpage
\clearpage
\begin{appendix}
% \section{Introduction}
\section{Annotation Guideline}
\label{sec:annotation}

\subsection{Entity Category}

 \begin{itemize}
\item \textbf{Task}: Applications, problems to solve, systems to construct.

E.g. information extraction, machine reading system, image segmentation, etc.

\item \textbf{Method}: Methods , models, systems to use, or tools, components of a system, frameworks. 

E.g. language model, CORENLP, POS parser, kernel method, etc.

\item \textbf{Evaluation Metric}: Metrics, measures, or entities that can express quality of a system/method. 

E.g. F1, BLEU, Precision, Recall, ROC curve, mean reciprocal rank, mean-squared error, robustness, time complexity, etc.

\item \textbf{Material}: Data, datasets, resources, Corpus, Knowledge base.

E.g. image data, speech data, stereo images, bilingual dictionary, paraphrased questions, CoNLL, Panntreebank, WordNet, Wikipedia, etc.

\item \textbf{Evaluation Metric}: Metric， measure， or term that can express quality of a system/method.

E.g. F1, BLEU, Precision, Recall, ROC curve, mean reciprocal rank, mean-squared error,robustness, compile time, time complexity...

\item \textbf{Generic}: General terms or pronouns that may refer to a entity but are not themselves informative, often used as connection words.

E.g model, approach, prior knowledge, them, it...

\end{itemize}

\subsection{Relation Category}
Relation link can not go beyond sentence boundary. We define 4 asymmetric relation types (\textit{Used-for}, \textit{Feature-of}, \textit{Hyponym-of}, \textit{Part-of}), together with 2 symmetric relation types (\textit{Compare}, \textit{Conjunction}).
\blue{B} always points to \red{A} for asymmetric relations

 \begin{itemize}
\item \textbf{Used-for}: \blue{B} is used for \red{A}, \blue{B} models \red{A}, \red{A} is trained on \blue{B}, \blue{B} exploits \red{A}, \red{A} is based on \blue{B}. E.g.
\begin{itemize}[label={}]
         \item The \blue{TISPER system} has been designed to enable many \red{text applications}.
         \item Our \blue{method} models \red{user proficiency}.
         \item Our \blue{algorithms} exploits \red{local soothness}.
\end{itemize}
\end{itemize}

\begin{itemize}
\item \textbf{Feature-of}: \blue{B} belongs to \red{A}, \blue{B} is a feature of \red{A}, \blue{B} is under \red{A} domain. E.g.
\begin{itemize}[label={}]
         \item \blue{prior knowledge} of the \red{model} 
         \item \blue{genre-specific regularities}  of  \red{discourse structure}
         \item \red{English text} in \blue{science domain}
\end{itemize}
\end{itemize}

\begin{itemize}
\item \textbf{Hyponym-of}: \blue{B} is a hyponym of \red{A}, \blue{B} is a type of \red{A}. E.g.
\begin{itemize}[label={}]
         \item \blue{TUIT} is a \red{software library}
         \item \red{NLP applications} such as \blue{machine translation} and \blue{language generation}
\end{itemize}
\end{itemize}

\begin{itemize}
\item \textbf{Part-of}: \blue{B} is a part of \red{A}... E.g.
\begin{itemize}[label={}]
         \item The \red{system} includes two models: \blue{speech recognition} and \blue{natural language understanding}
         \item We incorporate \blue{NLU module} to the \red{system}.
\end{itemize}
\end{itemize}

\begin{itemize}
\item \textbf{Compare}: Symmetric relation (use blue to denote entity). Opposite of conjunction, compare two models/methods, or listing two opposing entities. E.g.
\begin{itemize}[label={}]
         \item Unlike the \blue{quantitative prior}, the \blue{qualitative prior} is often ignored...
         \item We compare our \blue{system} with previous \blue{sequential tagging systems}...
\end{itemize}
\end{itemize}

\begin{itemize}
\item \textbf{Conjunction}: Symmetric relation (use blue to denote entity). Function as similar role or use/incorporate with. E.g.
\begin{itemize}[label={}]
         \item obtained from \blue{human expert} or \blue{knowledge base}
         \item NLP applications such as \blue{machine translation} and \blue{language generation}
\end{itemize}
\end{itemize}

\subsection{Coreference}
Two Entities that points to the same concept.
\begin{itemize}
\item \textbf{Anaphora and Cataphora}: 
\begin{itemize}[label={}]
         \item We introduce a \blue{machine reading system}... The \blue{system}...
         \item The \blue{prior knowledge} include...Such \blue{knowledge} can be applied to...
\end{itemize}
\item \textbf{Coreferring noun phrase}: 
\begin{itemize}[label={}]
         \item We develop a \blue{part-of-speech tagging system}...The \blue{POS tagger}...

\end{itemize}
\end{itemize}

\subsection{Notes}

\begin{enumerate}
\item Entity boundary annotation follows the ACL RD-TEC Annotation Guideline \cite{qasemizadeh2016acl}, with the extention that spans can be embedded in longer spans, only if the shorter span is involved in a relation.

\item Do not include determinators (such as ‘the’, ‘a’), or adjective pronouns (such as ‘this’,’its’, ‘these’, ‘such’) to the span. If generic phrases are not involved in a relation, do not tag them.

\item Do not tag relation if one entity is:
\begin{itemize}[]
\item Variable bound:

We introduce a neural based approach.. \textit{Its} benefit is...
\item The word \textit{which}:

We introduce a neural based approach, \textit{which} is a...
\end{itemize}
\item Do not tag coreference if the entity is

\begin{itemize}[]
\item Generically-used Other-ScientificTerm:

...advantage gained from \textit{local smoothness} which... We present algorithms exploiting \textit{local smoothness} in more aggressive ways...
\item Same scientific term but refer to different examples:

We use a \textit{data structure}, we also use another \textit{data structure}...
\end{itemize}
\item Do not label negative relations:

X is not used in Y or X is hard to be applied in Y

\end{enumerate}

\section{Annotation and Knowledge Graph Examples}
\label{sec:appendix}
\begin{figure*}[t]
\centering
\includegraphics[width=\textwidth, keepaspectratio,clip]{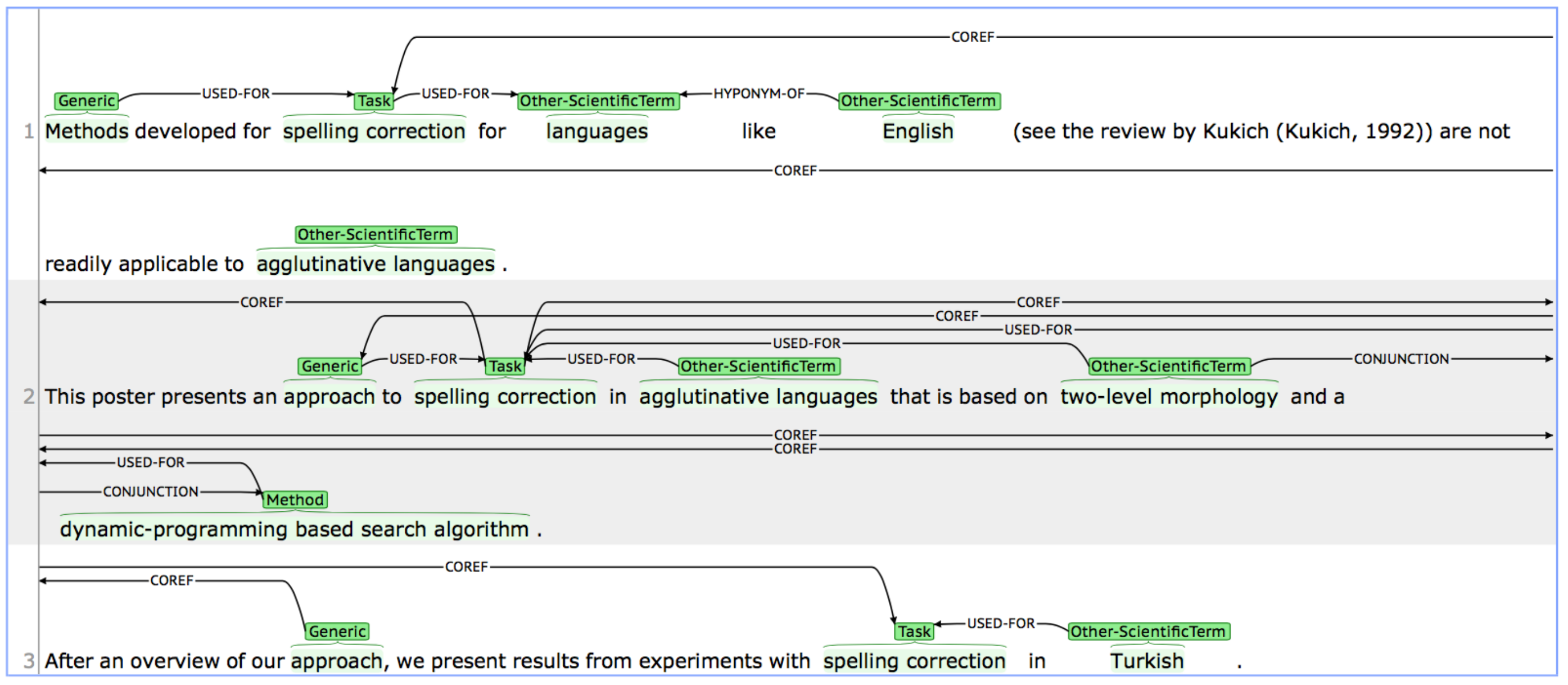}
\vspace{-1em}
\caption{
Annotation example 1 from ACL 
}\label{fig:ACL}
\vspace{-1em}
\end{figure*}

\begin{figure*}[t]
\centering
\includegraphics[width=0.9\textwidth, keepaspectratio,clip]{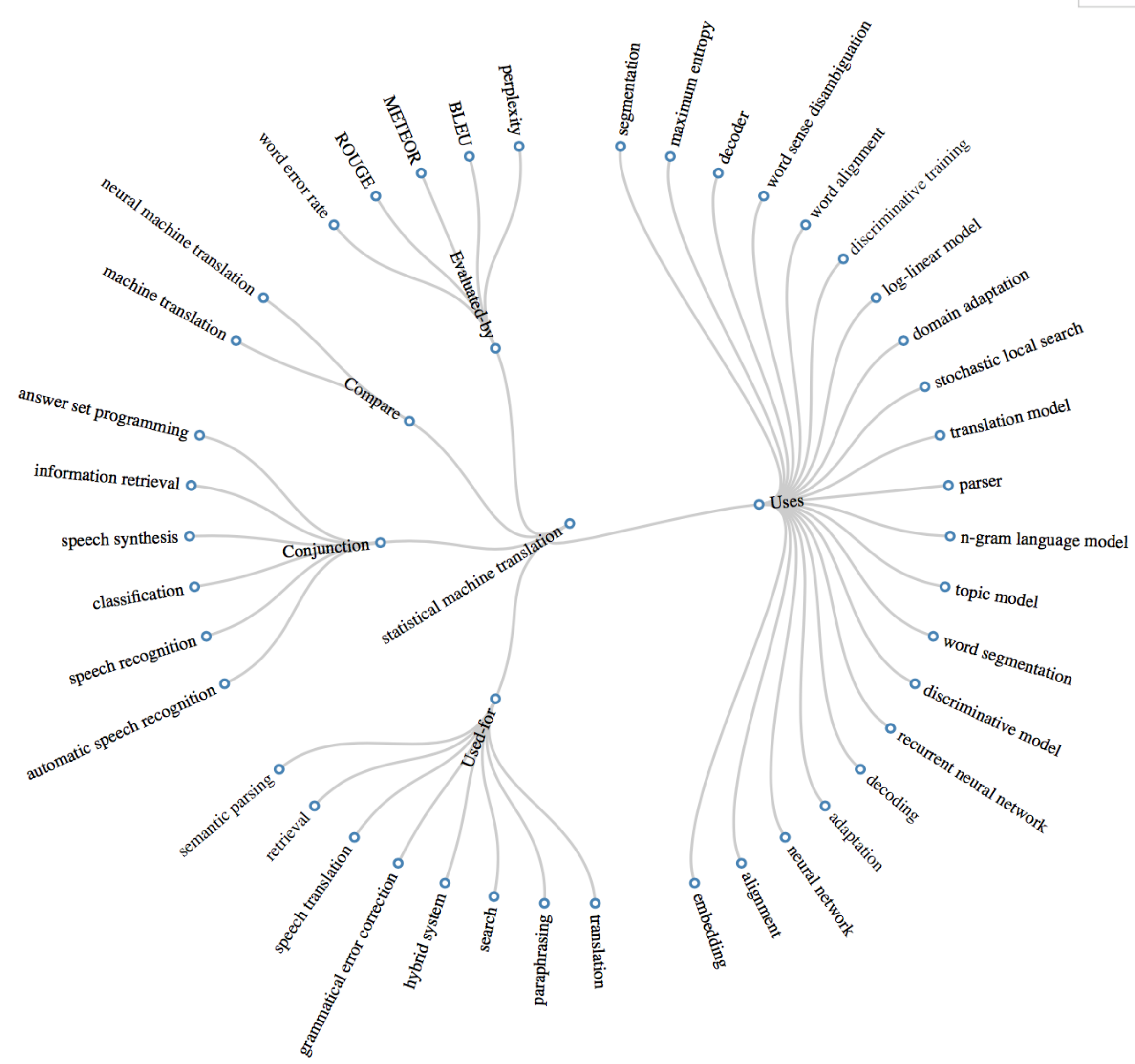}
\caption{
An example of our automatically generated knowledge graph centered on \textit{statistical machine translation}. This is the original figure of Figure \ref{fig:KG}.
}\label{fig:SMT}
\end{figure*}

Here we take a screen shot of the BRAT interface for an ACL paper in Figure~\ref{fig:ACL}.
% Following the annotation examples, 
We also attach the original figure of Figure 3 in Figure ~\ref{fig:SMT}. More examples can be found in the project website\footnote{\url{http://nlp.cs.washington.edu/sciIE/}}.

% \begin{figure*}[t]
% \centering
% \includegraphics[width=\textwidth, keepaspectratio,clip]{sfigures/CVPR_example.pdf}
% \vspace{-1em}
% \caption{
% Annotation example 2 from CVPR
% }\label{fig:CVPR}
% \vspace{-1em}
% \end{figure*}

% \begin{figure*}[t]
% \centering
% \includegraphics[width=\textwidth, keepaspectratio,clip]{sfigures/ICASSP.pdf}
% \vspace{-1em}
% \caption{
% Annotation example 3 from ICASSP
% }\label{fig:ICASSP}
% \vspace{-1em}
% \end{figure*}

% \newpage
% \begin{figure*}[t]
% \centering
% \includegraphics[width=\textwidth, keepaspectratio,clip]{sfigures/ICML.pdf}
% \vspace{-1em}
% \caption{
% Annotation example 4 from ICML
% }\label{fig:ICML}
% \vspace{-1em}
% \end{figure*}

% \begin{figure*}[t]
% \centering
% \includegraphics[width=\textwidth, keepaspectratio,clip]{sfigures/IJCAI.pdf}
% \vspace{-1em}
% \caption{
% Annotation example 4 from IJCAI
% }\label{fig:IJCAI}
% \vspace{-1em}
% \end{figure*}

\newpage

% \begin{figure*}[t]
% \centering
% \includegraphics[width=\textwidth, keepaspectratio,clip]{sfigures/CNN.pdf}
% \vspace{-1em}
% \caption{
% Automatically generated knowledge graph example centered on \textit{convolutional neural network} 
% }\label{fig:CNN}
% \vspace{-1em}
% \end{figure*}

% \begin{figure*}[t]
% \centering
% \includegraphics[width=\textwidth, keepaspectratio,clip]{sfigures/CRF.pdf}
% \vspace{-1em}
% \caption{
% Automatically generated knowledge graph example centered on \textit{conditional random field} 
% }\label{fig:CRF}
% \vspace{-1em}
% \end{figure*}

% \begin{figure*}[t]
% \centering
% \includegraphics[width=\textwidth, keepaspectratio,clip]{sfigures/NER.pdf}
% \vspace{-1em}
% \caption{
% Automatically generated knowledge graph example centered on \textit{named entity recognition} 
% }\label{fig:NER}
% \vspace{-1em}
% \end{figure*}

% \begin{figure*}[t]
% \centering
% \includegraphics[width=\textwidth, keepaspectratio,clip]{sfigures/imagesegmentation.pdf}
% \vspace{-1em}
% \caption{
% Automatically generated knowledge graph example centered on \textit{image segmentation} 
% }\label{fig:IS}
% \vspace{-1em}
% \end{figure*}

\end{appendix}

\end{document}